\documentclass[10pt,letterpaper]{article}

\usepackage{cogsci}

\cogscifinalcopy 

\usepackage{pslatex}
\usepackage{apacite}
\usepackage{float} 

\usepackage{amssymb}            
\usepackage{mathtools}          
\usepackage{mathrsfs}           
\usepackage{graphicx}           
\usepackage{subcaption}         
\usepackage[space]{grffile}     
\usepackage{url}                

\usepackage{algorithm}     
\usepackage{algpseudocode}

\algrenewcommand\algorithmicrequire{\textbf{Input:}}
\algrenewcommand\algorithmicensure{\textbf{Output:}}

\usepackage{amsthm}
\theoremstyle{definition}
\newtheorem*{definition*}{Definition}

\title{Learning telic-controllable state representations}
 
\author{{\large \bf Nadav Amir (nadav.amir@princeton.edu)} \\
  Princeton Neuroscience Institute, Princeton University \\
  Princeton, NJ 08540 USA
  \AND {\large \bf Stas Tiomkin (stas.tiomkin@ttu.edu)} \\
  Department of Computer Science, Texas Tech University \\
  Lubbock, TX 79409 USA}
\begin{document}

\maketitle

\begin{abstract}
Computational models of purposeful behavior comprise both descriptive and prescriptive aspects, used respectively to ascertain and evaluate situations in the world. In reinforcement learning, prescriptive reward functions are assumed to depend on predefined and fixed descriptive state representations. Alternatively, these two aspects may emerge interdependently: goals can shape the acquired state representations and vice versa. Here, we present a computational framework for state representation learning in bounded agents, where descriptive and prescriptive aspects are coupled through the notion of goal-directed, or telic, states. We introduce the concept of telic-controllability to characterize the tradeoff between the granularity of a telic state representation and the policy complexity required to reach all telic states. We propose an algorithm for learning telic-controllable state representations, illustrating it using a simulated navigation task. Our framework highlights the role of deliberate ignorance -- knowing what to ignore -- for learning state representations that balance goal flexibility and cognitive complexity. 
\end{abstract}

\begin{figure}[htbp]
    \centering
    \includegraphics[width=0.45\textwidth]{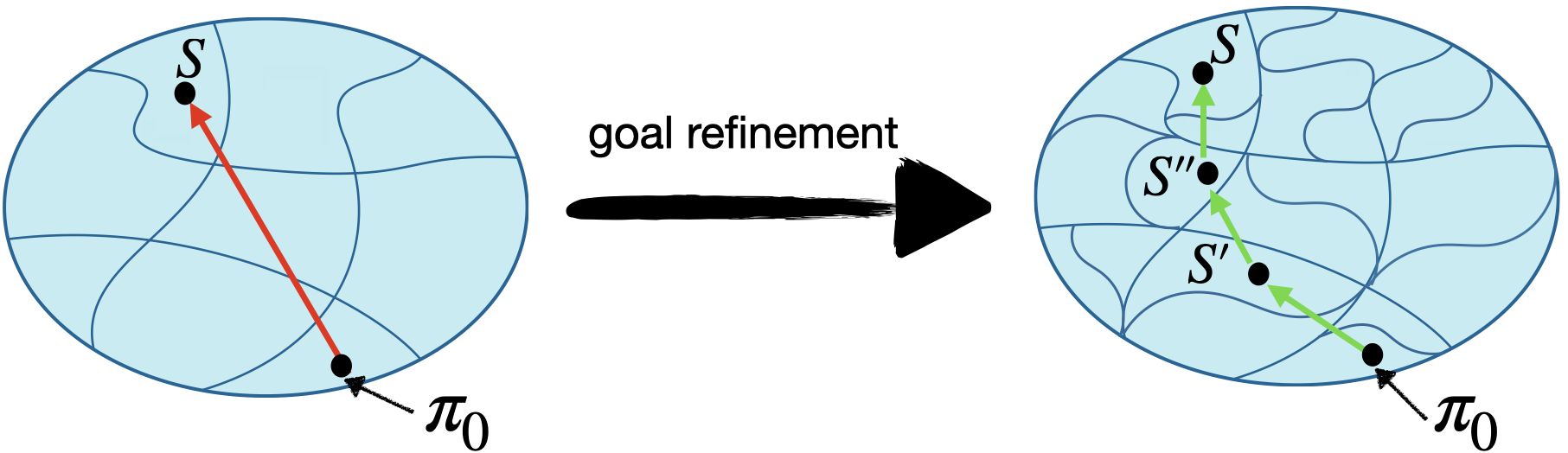}
    \caption{\small \textbf{The granularity-complexity tradeoff:} within the framework developed in this paper, state representations partition all experiences (blue ellipses) into preference-ordered classes called ``telic states'' (curved regions), each consisting of all experience distributions that are approximately equivalent with respect to the agent's goal. \textbf{Left:} an agent with a coarse-grained goal is unable to reach a desired telic state, $S$, which is too distant, in a statistical sense, from its default policy $\pi_0$. \textbf{Right:} refining the goals yields a state representation that is more controllable since all telic states can be reached using incremental policy update steps.}
    \label{fig:goal_refinement_cartoon}
\end{figure}

\section{Introduction}
    How do goals shape the way learning agents represent their experience? This fundamental question has only recently started drawing increased attention in both cognitive science and AI \cite{molinaro2023goal,muhle2023goal, radulescu2019holistic,eysenbach2022contrastive,florensa2018automatic, wang2024goal}. For example, recent empirical work suggest that humans may structure their experience into discrete states by balancing the utility and complexity of their representations \cite{fang2025humans}. However, it remains unclear, from a theoretical perspective, how should computationally bounded learning agents adjust their state representations when their goals are changing. For example, consider a rodent navigating a complex maze with changing reward contingencies \cite{krausz2023dual}, or a robot trained to do various object manipulation tasks with sparse rewards \cite{andrychowicz2017hindsight}. How should such learning agents represent their tasks in ways that facilitate adaptation to shifting goals using limited computational resources? Prior works have addressed the problem of efficient state representation learning using bisimulation  \cite{zhang2020learning, wang2024building} or option-based methods \cite{abel2020value}. While these approaches can provide efficient heuristics for state abstraction in Markovian settings, our aim here is to characterize the fundamental tradeoff between the granularity of a state representation (which may or may not be Markovian), and the computational resources needed to generate a policy that can efficiently utilize it. We present a principled approach to this problem, leveraging a recently proposed theoretical framework of goal-directed, or \emph{telic}, state representation learning \cite{amir2023states}. We define a novel property, \emph{telic-controllability}, characterizing the ability to reach all states within a given telic state representation using complexity bound policies. We describe a telic state representation learning algorithm and illustrate it using a simple navigation task by showing how complexity bounded agents can learn a telic-controllable state representation that can adapt to shifting goals.

\section{Formal setting}
\subsection{Telic states as goal-equivalent experiences}
We assume the setting of a perception-action cycle, i.e., sequences of observation-action pairs representing the flow of information between agent and environment. We denote by $\mathcal{O}$ and $\mathcal{A}$ the set of possible observations and actions, respectively. An experience sequence, or \emph{experience} for short, is a finite sequence of observation-action pairs: $h=o_1,a_1,o_2,a_2,...,o_n,a_n$. For every non-negative integer, $n\geq0$, we denote by $\mathcal{H}_n\equiv(\mathcal{O}\times\mathcal{A})^n$ the set of all experiences of length $n$. The collection of all finite experiences is denoted by $\mathcal{H}=\cup_{n=1}^{\infty}\mathcal{H}_n$. In non-deterministic settings, it will be useful to consider distributions over experiences rather than individual experiences themselves and we denote the set of all probability distributions over finite experiences by $\Delta(\mathcal{H})$. 
Following Bowling et~al.~\citeyear{bowling2022settling}, we define a \emph{goal} as a binary preference relation over experience distributions. For any pair of experience distributions, $A,B\in\Delta(\mathcal{H})$, we write $A\succeq_gB$ to indicate that experience distribution $A$ is weakly preferred by the agent over $B$, i.e., that $A$ is at least as desirable as $B$, with respect to goal $g$. 
When $A\succeq_gB$ and $B\succeq_gA$ both hold, $A$ and $B$ are equally preferred with respect to $g$, denoted as $A\sim_gB$. We observe that $\sim_g$ 
is an equivalence relation, i.e., it satisfies the following three properties, for any $A,B,C\in\Delta(\mathcal{H})$:
\begin{itemize}
    \item Reflexivity: $A\sim_gA$ for all $A\in\Delta(\mathcal{H}).$
    \item Symmetry: $A\sim_gB$ implies $B\sim_gA$ for all $A,B\in\Delta(\mathcal{H}).$
    \item Transitivity: if $A\sim_gB$ and $B\sim_g C$ then $A\sim_g C$ for all $A,B,C\in\Delta(\mathcal{H}).$
\end{itemize}
Therefore, every goal induces a partition of $\Delta(\mathcal{H})$ into disjoint sets of equally desirable experience distributions. For goal $g$, we define the goal-directed, or \emph{telic}, state representation, $\mathcal{S}_g$, as the partition of experience distributions into equivalence classes it induces: $\mathcal{S}_g= \Delta(\mathcal{H})/\sim_g$.
In other words, each telic state represents a generalization over all equally desirable experience distributions. This definition captures the intuition that agents need not distinguish between experiences that are equivalent, in a statistical sense, with respect to their goal. Furthermore, since different telic states are, by definition, non-equivalent with respect to $\succeq_g$, the goal $g$ also determines whether a transition between any two telic states brings the agent in closer alignment to, or further away from its goal. 

\subsection{Learning with telic states}
 \label{sec:learning_with_telic_states}
How can telic state representations guide goal-directed behavior? To address this question, we recall the definition of a \emph{policy}, $\pi$, as a distribution over actions given the past experience sequence and current observation:
\begin{equation}
\label{eq:policy_def_orig_appendix_2}
\pi(a_i|o_1,a_1,...,o_i).
\end{equation}
Analogously, we can define an \emph{environment}, $e$, as a distribution over observations given the past experience sequence: 
\begin{equation}
e(o_i|o_1,a_1,...,a_{i-1}).
\end{equation}
The distribution over experience sequences can be factored, using the chain rule, as follows: 
\begin{equation}
\begin{split}
    &P_\pi(o_1,a_1,...,o_n,a_n)=P(o_1,a_1,...,o_n,a_n|e,\pi)=\\
    &\prod_{i=1}^n e(o_i|o_1,a_1,...,a_{i-1})\pi(a_i|o_1,a_1,...,o_i).
\end{split}
\end{equation} 
Typically, the environment is assumed to be fixed, and hence not explicitly parameterized in $P_\pi(h)$ above.
The definition of telic states as goal-induced equivalence classes can now be extended to equivalence classes of policy-induced experience distributions as follows: 
\begin{equation}
    \pi_1\sim_g\pi_2\iff P_{\pi_1}\sim_gP_{\pi_2}.
\end{equation}
The question we are interested in here is the following: how can an agent learn an efficient policy for reaching a desired telic state? In other words, how can an agent acquire policies that are likely to generate experiences belonging to a certain telic state $S_i\in\mathcal{S}_g$? 
To answer this, we consider the empirical distribution of $N$ experience sequences generated by policy $\pi$:
\begin{equation}
    \hat{P}_\pi(h)=\frac{|\{i:h_i=h\}|}{N}.
\end{equation}
By Sanov's theorem \cite{cover1999elements}, the likelihood that $ \hat{P}_\pi(h)$ belongs to telic state $S_i$ decays exponentially with a rate of 
\begin{equation}
\label{eq:telic_distance}
R=D_{KL}(P_i^{\star}||P_\pi)
\end{equation}
where,
\begin{equation}
\label{eq:info-proj}
P_i^\star = arg\min_{P\in S_i} D_{KL}(P||P_\pi),
\end{equation} is the \emph{information projection} of $P_\pi$ onto $S_i$, i.e., the distribution in $S_i$ which is closest, in the KL sense, to $P_\pi$. Thus, $R$ can be thought of as the ``telic distance'' from $\pi$ to $S_i$ since it determines the likelihood that experiences sampled from $P_\pi$ belong to the telic state $S_i$. Assuming a  policy parameterized by $\theta$, the following policy gradient method updates $\pi_\theta$ in a way that minimizes its telic distance to $S_i$:
\begin{equation}
    \label{eq:policy_grad_step}
    \theta_{t+1}=\theta_t-\eta\nabla_\theta D_{KL}(P_i^\star||P_{\pi_\theta}),
\end{equation}
where $\eta>0$ is a learning rate parameter.

\subsection{Experience features and discrimination sensitivity}
\label{sec:sensitivity}
A natural way of representing goals, i.e., preferences over experience distributions, is by comparing the likelihood that experiences generated from different distributions will belong to some subset $\Phi_g\subset\mathcal{H}$ representing some desired property of experiences. For example, for the goal of solving a maze, $\Phi_g$ might be the set of all experiences, i.e., path trajectories, that reach the exit. Formally, for two experience distributions, $A$ and $B$, the agent will prefer the one that is more likely to generate
experiences belonging to $\Phi_g$:
\begin{equation*}
    \label{eq:pref_by_property}
    A\succeq_gB : \sum_{h\in\Phi_g}A(h)\geq \sum_{h\in\Phi_g}B(h).
\end{equation*}
The sensitivity parameter, $\epsilon$, effectively determining the maximum difference, in terms of desirable outcome likelihoods, that the agent is willing to ignore in order to reduce representational complexity. In the maze example, experience distribution $A$ would be preferred over $B$ if it is more likely to generate trajectories that reach the exit. 
Importantly, Eq.~\ref{eq:pref_by_property} implies that $A$ and $B$ are equivalent only when $\sum_{h\in\Phi_g}A(h)$ and $\sum_{h\in\Phi_g}B(h)$ are precisely equal, which is unlikely in realistic, noisy environments. A more reasonable assumption is that agents can discriminate sampling likelihoods at some finite sensitivity level, $\epsilon>0$, such that: 
\begin{equation}
\label{eq:equiv_granularity}
A\sim^{(\epsilon)}_g B \iff |\sum_{h\in\Phi_g}A(h)-\sum_{h\in\Phi_g}B(h)|\leq\epsilon. 
\end{equation}
In the maze example, this means that two trajectory distributions are considered equivalent if their respective likelihoods of generating exit-reaching trajectories are within $\epsilon$ of each other.   
As we shall see in the following sections, the discrimination sensitivity parameter, $\epsilon$, controls the tradeoff between the granularity of a telic state representation and the policy complexity needed to reach all telic states. 

\subsection{Telic-controllability}
\label{sec:telic_controllability}
In this section, we introduce the notion of \emph{telic-controllability}, a joint property of an agent and a telic state representation, that characterizes whether or not the agent is able to reach all possible telic states using complexity-limited policy update steps. Towards this, we first define an agent's \emph{policy}, $\pi$, as a distribution over actions given the past experience sequence and current observation: $\pi(a_i|o_1,a_1,...,o_i)$.
Assuming a fixed environment, the definition of telic states as goal-induced equivalence classes induces corresponding equivalence classes of policy-induced experience distributions as follows: 
\begin{equation}
    \label{eq:policy_equivalence}
    \pi_1\sim_g\pi_2\iff P_{\pi_1}\sim_gP_{\pi_2}.
\end{equation}
As detailed above, this mapping between policies and telic states provides a unified account of goal-directed learning in terms of the statistical distance between policy-induced distributions and desired telic states. To explore this notion, we introduce a new property -- telic-controllability -- that plays a central role in the following sections. A representation is called telic-controllable if any state can be reached using a finite number, $N$, of complexity-limited policy updates, starting from the agent's default policy, $\pi_0$, where the complexity of a policy update step is quantified by the Kullback-Leibler (KL) divergence between the post and pre-update step policies. Formally, we have the following:
\begin{definition*}[telic-controllability]
A telic-state representation, $\mathcal{S}_g$, induced by the goal, $g$, is \emph{telic-controllable} with respect to a default policy, $\pi_0$, and a policy complexity capacity, $\delta\ge0$, if the following holds:
\begin{equation}
\label{eq:telic_controllability}
\begin{split}
&\forall S\in \mathcal{S}_g\; \exists \{\pi_t,S_t\}_{t=0}^N,N>0  \text{ s.t. } \forall t<N\\
&\big(S_t=[P_{\pi_t}]_{\sim_g}\big) \wedge \big(D_{KL}(P_{\pi_{t+1}}||P_{\pi_{t}})\leq \delta\big)\wedge \big(S_N=S\big),
\end{split}
\end{equation}
\end{definition*}
where $[P_{\pi_t}]_{\sim_g}$ is the goal-induced equivalence class, i.e., telic state, containing $P_{\pi_t}$. This definition generalizes the familiar control theoretic notion of controllability in two important ways. First, it applies to telic states, i.e., classes of distributions over action-outcome trajectories, rather than by n-dimensional vectors -- the standard control theoretic setting. Second, it takes into account the complexity capacity limitations of the agent, using information theoretic quantifiers to constrain the maximal complexity of policy update steps an agent can take in attempting to reach one telic state from another. As illustrated in the next section, telic-controllability is a desirable property since it means that agents can flexibly adjust to shifting goals using bounded policy complexity resources.
 
 \subsection{State representation learning algorithm}
 \label{sec:learning_algorithm}
 A central feature of our approach is the duality it establishes between goals and state representations. In this section, we utilize this duality to develop an algorithm for learning a telic-controllable state representation, or, equivalently, finding a goal that produces such a state representation. The algorithm receives as inputs the agent's current goal, $g$ (represented, e.g., by an ordered set of desired experience features), and default policy, $\pi_0$, along with its policy complexity capacity, $\delta$, and the discrimination sensitivity parameter $\epsilon$. Its output consists of a new goal $g'$ such that $\mathcal{S}_{g'}$ is telic controllable with respect to $\pi_0$ and $\delta$. 
 The main idea is to split any unreachable telic state, $S$, i.e., one that cannot be reached from $\pi_0$ using policy update steps with complexity less than $\delta$. State splitting is accomplished by generating a new, intermediate, telic state, $S_M$, lying between the agent's default policy induced distribution, $P_{\pi_0}$, and its information projection on the unreachable telic state, i.e., the distribution $P^*\in S$ that is closest to $P_{\pi_0}$, in the KL sense. The intermediate telic state, $S_M$, is then defined as the set of all distributions that are $\epsilon$-equivalent to $P_M$ (Eq.~\ref{eq:equiv_granularity}), where $P_M$ is the convex combination of $P^*$ and $P_{\pi_0}$ lying at a KL distance of $\delta$ from $P_{\pi_0}$.  After generating the new state, $S_M$, the goal is updated to reflect the proper ordering between the default policy state $S_0$, the intermediate state $S_M$, and the originally unreachable state $S$, such that elements of $S_M$ are between $S_0$ and $S$ in terms of preference. Pseudocode for the learning algorithm is provided in Algorithm~\ref{alg:learn_telic_sr}. The algorithm makes use of an auxiliary procedure, \textproc{FindReachableStates} (Algorithm~\ref{alg:get_reachable_states}), to find all reachable states, given the agent's goal, $g$, default policy, $\pi_0$, and policy complexity constraint, $\delta$. This auxiliary procedure performs a recursive search, similar to depth-first search methods, attempting to find policies that are closest, in the KL sense, to currently unreachable telic states, while still sufficiently close to the agent's current policy, as not to exceed the policy complexity capacity. Its main optimization step (line 3) can be implemented, e.g., using policy gradient over the information projection of $P_{\pi_0}$ on $S$. 
\begin{algorithm}[ht]
\caption{Telic-controllable state representation learning}\label{alg:learn_telic_sr}
\begin{algorithmic}[1]
\Require{$\pi_0 \text{: default policy, } g \text{: current goal, } \newline \delta \text{: policy complexity capacity, } \epsilon \text{: sensitivity.}$}
\Ensure $g' \text{: new goal such that } \mathcal{S}_{g'}$ is telic-controllable with respect to $\pi_0$ and $\delta$
\State $\mathcal{R} \gets [P_{\pi_0}]_{\sim_{g}}$ \Comment{initialize reachable state set}
\State $g'\gets g$  \Comment{initialize new goal}
\While {$\mathcal{R} \ne \mathcal{S}_{g'}$}
    \State $\mathcal{R} \gets \Call{FindReachableStates}{\pi_0,g',\delta}$ \Comment{see algorithm~\ref{alg:get_reachable_states} below}
    \For{$S\in \mathcal{S}_{g'}\setminus\mathcal{R}$} \Comment{for each unreachable state}
        \State $P^{*} \gets \arg\min_{P\in S} D_{KL}(P||P_{\pi_{0}})$ \Comment{information projection of $P_{\pi_0}$ on $S$}

       \State $M = \arg\max_{t\in[0,1]} t \text{ s.t. } D_{KL}\big((t P^{*}+(1-t)P_{\pi_{0}})||P_{\pi_0}\big)\leq\delta$ 
        
        \State $P_M = M P^{*}+(1-M)P_{\pi_{0}}$ \Comment{convex combination of $P^*$ and $P_{\pi_0}$}
        \State $S_M \gets \{P : P\sim^{(\epsilon)}_g P_M\}$ \Comment{$\epsilon$-neighborhood of $P_M$}
        \If{$P_{\pi_{0}}\leq_gP^{*}$} \Comment{update goal with preference order for $S_M$}
            \State $g'\gets g' \cup \{(p,q)_{\leq_{g'}}\in S_M\times S \}\cup \{(r,p)_{\leq_{g'}}\in S_0\times S_M\}$
        \ElsIf{$P^{*}\leq_gP_{\pi_{0}}$}
            \State $g'\gets g' \cup \{(q,p)_{\leq_{g'}}\in S\times S_M \}\cup \{(p,r)_{\leq_{g'}}\in S_M\times S_0 \}$
        \EndIf
    \EndFor
\EndWhile
\State \Return{$g'$}
\end{algorithmic}
\end{algorithm}

\begin{algorithm}
\caption{Finding reachable states }\label{alg:get_reachable_states}
\begin{algorithmic}[1]
\Require{$\pi_0 \text{: initial policy, } g \text{: goal, } \newline \delta \text{: policy complexity constraint.}$}
\Ensure{all telic states in $\mathcal{S}_g$ reachable from ${\pi_0}$ by $\delta$-complexity limited policy update steps}
\Procedure{RecursiveReach}{$\pi,g,\delta,\mathcal{R}$}
    \For{$S\in \mathcal{S}_g\setminus\mathcal{R}$} \Comment{for every unreached state $S$}
        \State $\pi_\theta\gets\arg\min_\theta D_{KL}(S||P_{\pi_{\theta}}) \text{ s.t. } D_{KL}(P_{\pi_{\theta}}||P_{\pi})\leq\delta$ \Comment{optimize policy to reach $S$}
        \If{$[P_{\pi_\theta}]_{\sim_g}\notin\mathcal{R}$} \Comment{if new state reached}
         \State $\mathcal{R}\gets \mathcal{R}\cup[P_{\pi}]_{\sim_g}$ \Comment{add current state to reachable set}
        \State $\mathcal{R} \gets \Call{RecursiveReach} {\pi_\theta,g,\delta,\mathcal{R}}$  \Comment{continue from current state}
        \EndIf
    \EndFor
    \State \Return{$\mathcal{R}$}
\EndProcedure
\Procedure{FindReachableStates}{$\pi_0, g, \delta$}
    \State $\mathcal{R}_0 \gets [P_{\pi_0}]_{\sim_g}$ \Comment{initialize reachable set} 
    \State $\mathcal{R} \gets$ \Call{RecursiveReach} {$\pi_0,g,\delta,\mathcal{R}_0$} \Comment{try to reach all states recursively}
    \State \Return{$\mathcal{R}$} \Comment{return set of reachable states}
\EndProcedure

\end{algorithmic}
\end{algorithm}

\section{Illustrative example: dual goal navigation task}
\label{sec:example}
\begin{figure}[ht]
    \centering
    \includegraphics[width=0.45\textwidth]{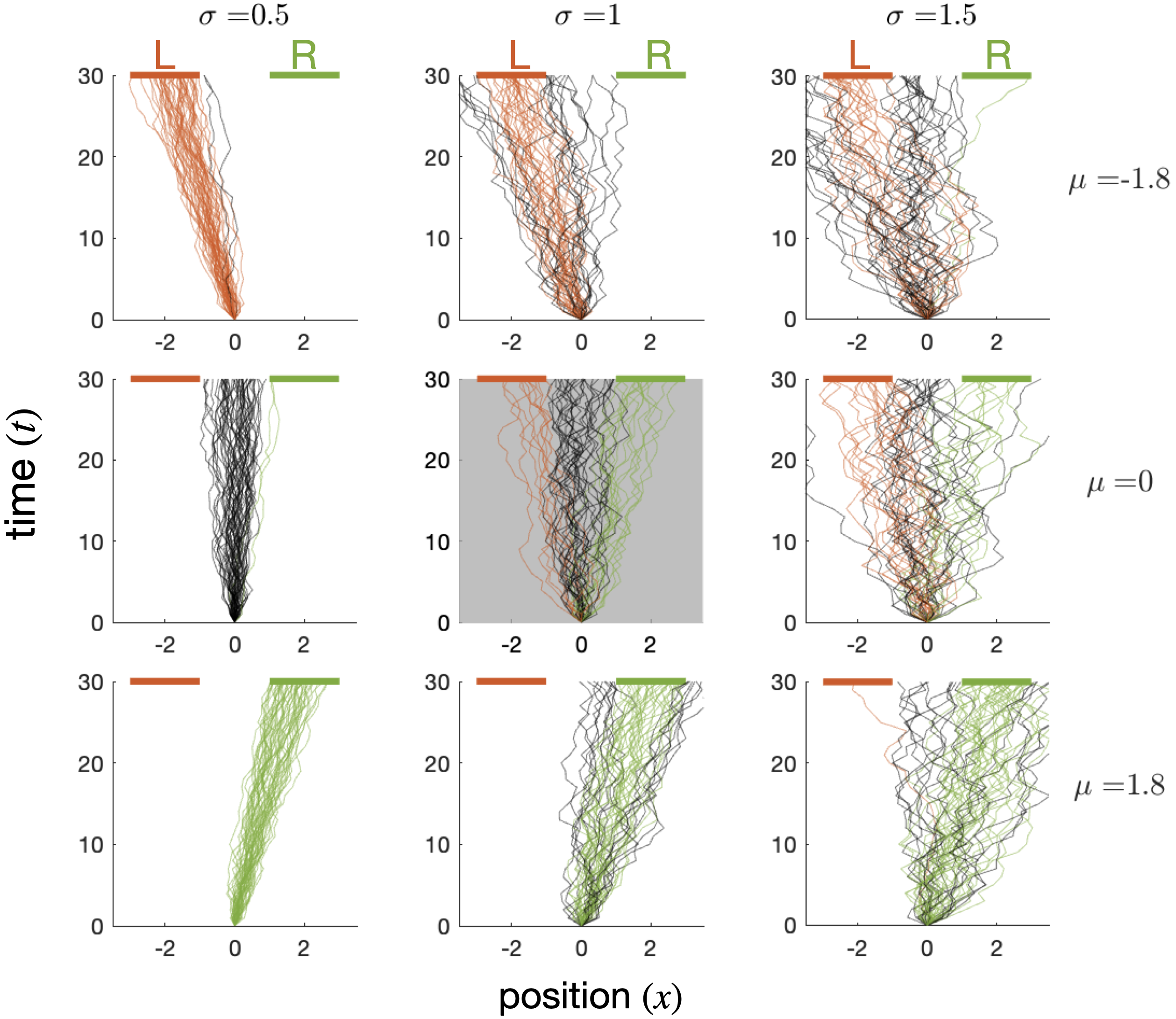}
    \caption{\small \textbf{Dual goal navigation task:} each tile shows 500 one-dimensional random walk trajectories of length $T=30$, generated by a Gaussian policy parameterized by the mean ($\mu$) and standard deviation ($\sigma$) of position update step (x-axis) across time (y-axis). Regions of interest $R$ and $L$ consist of line segments centered around $x_R=2$ and $x_L=-2$, shown as green and red lines respectively at $T=30$. Trajectories reaching one of the goals are plotted in the corresponding color, illustrating the relationship between policy parameters and goal reaching likelihoods. The default policy, $(\mu_0,\sigma_0)=(0,1)$, shown in the center gray tile, is equally likely to reach $R$ and $L$.}
    \label{fig:random_walk_traj}
\end{figure}

In this section, we illustrate the proposed state representation framework and learning algorithm using a simple navigation task in which an agent performs a one dimensional random walk, starting at location $x_0=0$, with the goal of reaching one of two non overlapping regions of interest after a fixed number, $T=30$, of steps. 
The agent's policy is defined as a stochastic mapping between its current and next position and is parameterized by the mean and standard deviation ($\mu$ and $\sigma$, respectively) of a Gaussian update step: $\pi(x_{t+1}|x_t;\mu,\sigma)=x_t+\eta_t,\quad \eta_t\sim\mathcal{N}(\mu,\sigma).$
For brevity, we denote by $\pi(\mu,\sigma)$  a policy with a $\mathcal{N}(\mu,\sigma)$ distributed noise term. A graphical illustration of the task and sample trajectories for different policies is shown in Fig.~\ref{fig:random_walk_traj}.
\par
\begin{figure}[ht]
        \centering
            \includegraphics[width=0.45\textwidth]{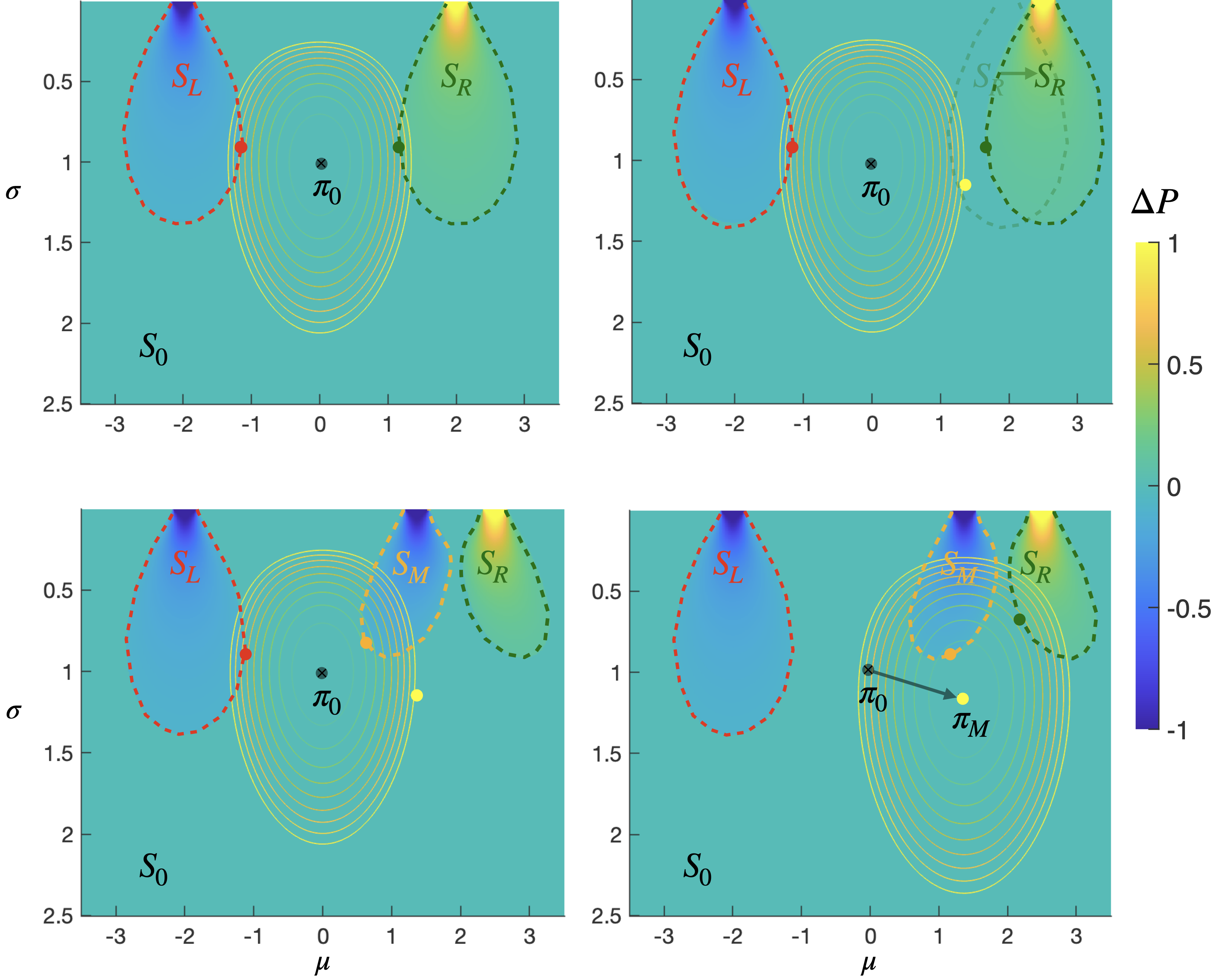} 
        \caption
        {\small \textbf{Telic state representation learning for navigation task with shifting goals:} points in $(\mu,\sigma)$ policy space colored by the difference between their probability of reaching unit length regions, $R$ and $L$, centered around $2$ and $-2$ respectively, at time $T=30$. \textbf{Top left:} telic states $S_L$ and $S_R$ (outlined by red and green dashed lines, respectively) consist of policies that are more likely to reach the corresponding region by a threshold of $\epsilon=0.1$ or more. Contour lines indicate isometric policy complexity levels, relative to the default policy $\pi_0:(\mu_0=0,\sigma_0=1)$ (black dot), for a capacity bound of $\delta=1$ bit. Green and red dots show the information projection of $\pi_0$ on $S_R$ and $S_L$ respectively, i.e., the policies each telic state closest to $\pi_0$ in KL-divergence~\textbf{Top right:} shifting the center of $R$ to $2.5$, renders $S_R$ unreachable from $\pi_0$ with $\delta$ bounded policy complexity. The policy $\pi_M:(\mu_M,\sigma_M)$ (yellow dot) is the one closest to $S_R$ while still within the complexity capacity of the agent.~\textbf{Bottom left:} splitting $S_R$ by inserting an intermediate telic-state, $S_M$, centered around $\mu_M$. By construction, the nearest distribution to $\pi_0$ in $S_M$, in the KL sense (orange dot), is within the agent's complexity capacity. ~\textbf{Bottom right:} both $S_M$ and $S_R$ are reachable with respect to the agent's new default policy, $\pi_M(\mu_M=1.37,\sigma_M=1.15)$ (see algorithm~\ref{alg:learn_telic_sr} for details); the new telic state representation $\{S_0,S_L,S_M,S_R\}$ is telic controllable with respect to $\pi_0(0,1),\,\delta=1,$ and $N=1$.} 
    \label{fig:phase_plots}
    \end{figure}

Since the sum of normally distributed variables is also normally distributed, a policy $\pi(\mu,\sigma)$ induces a Gaussian distribution over the final location of the agent:
\begin{equation}
\label{eq:random_walk_final_loc_dist}
p(x_T\mid x_0=0;\mu,\sigma)=\mathcal{N}(T\mu,\sqrt{T}\sigma).
\end{equation}
To account for goal-directed behavior, we define a right and a left region of interest, $R$ and $L$, consisting of unit radius segments centered around $x_R=2$ and $x_L=-2$ respectively. Thus, $R=[R_1,R_2]=[1,3]$ and $L=[L_1,L_2]=[-3,-1]$. For the purpose of this example, we assume that the agent wants to reach $R$ but avoid $L$, at time $T$. For example, for a rodent navigating a narrow corridor, $R$ and $L$ may indicate segments of the corridor where a reward (e.g., food) and a punishment (e.g., air puff) are administered, respectively.
We can express the agent's goal in terms of preferences over policies by defining $\Delta P(\mu,\sigma)=p(x_T\in R\mid \mu,\sigma)-p(x_T\in L\mid \mu,\sigma)$ as the difference between the probabilities that the agent will reach regions $R$ and $L$ at time $T$, with a policy $\pi(\mu,\sigma)$. The agent's goal can now be defined as a preference for policies with higher $\Delta P$ values. However, as explained above, due to the agent's finite discrimination resolution, it can only detect whether $\Delta P$ is above or below the sensitivity threshold, $\epsilon$. Thus, using Eq.~\ref{eq:policy_equivalence}, the agent's goal, $g$, can be expressed by the following preference relation over policies, where we denote, for brevity,  $\pi(\mu_i,\sigma_i)$ and $\Delta P(\mu_i,\sigma_i)$ as $\pi_i$ and $\Delta P_i$, respectively, for $i=1,2$:
\begin{equation}
\label{eq:pref_structure_R}
\begin{split} 
&\pi_1\succeq_g \pi_2\iff\big(\Delta P_1\geq\epsilon\geq\Delta P_2\big)\lor
\big(\Delta P_1\geq-\epsilon\geq\Delta P_2\big),
\end{split}
\end{equation}
where first term on the r.h.s. of Eq.~\ref{eq:pref_structure_R} captures the \emph{desirability} of $R$ -- the agent prefers policies that have a probability \emph{higher} than $\epsilon$ of reaching $R$ over ones that do not; while the second term captures the \emph{undesirability} of $L$ -- the agent prefers policies that have a probability \emph{lower} than $\epsilon$ to reach $L$ than ones that do not. We recall that telic states can be defined by policies that are similarly preferred, under the agent's discrimination threshold, $\epsilon$, which determines the borders between the resulting telic states. The telic state representation for the goal $g$ defined by Eq.~\ref{eq:pref_structure_R}, and a threshold parameter of $\epsilon=0.1$ is visualized in Fig.~\ref{fig:phase_plots} (top left). Telic state $S_R$ ($S_L$), is shown as a colored region bounded by a dotted green (red) line, consisting of all policies that are more (less) likely to reach $R$ than $L$ by a probability margin of $\epsilon$ or more. Policies that are roughly equally likely to reach $R$ or $L$, i.e., whose difference in $\Delta P$ is smaller than $\epsilon$, constitute an additional ``default'' telic state, $S_0$ (teal background), in which the agent is agnostic to which region is it more likely to reach. 
\begin{equation}
\label{eq:telic_states_def}
\begin{split}
&S_R=\left\{(\mu,\sigma)|\Delta P(\mu,\sigma)\geq \epsilon\right\},\\
&S_L=\left\{(\mu,\sigma)|\Delta P(\mu,\sigma)\leq -\epsilon\right\},\\
&S_0=\left\{(\mu,\sigma)||\Delta P(\mu,\sigma)|\leq \epsilon\right\}.
\end{split}
\end{equation}
    
\begin{figure}[ht]
\centering
    \includegraphics[width=0.45\textwidth]{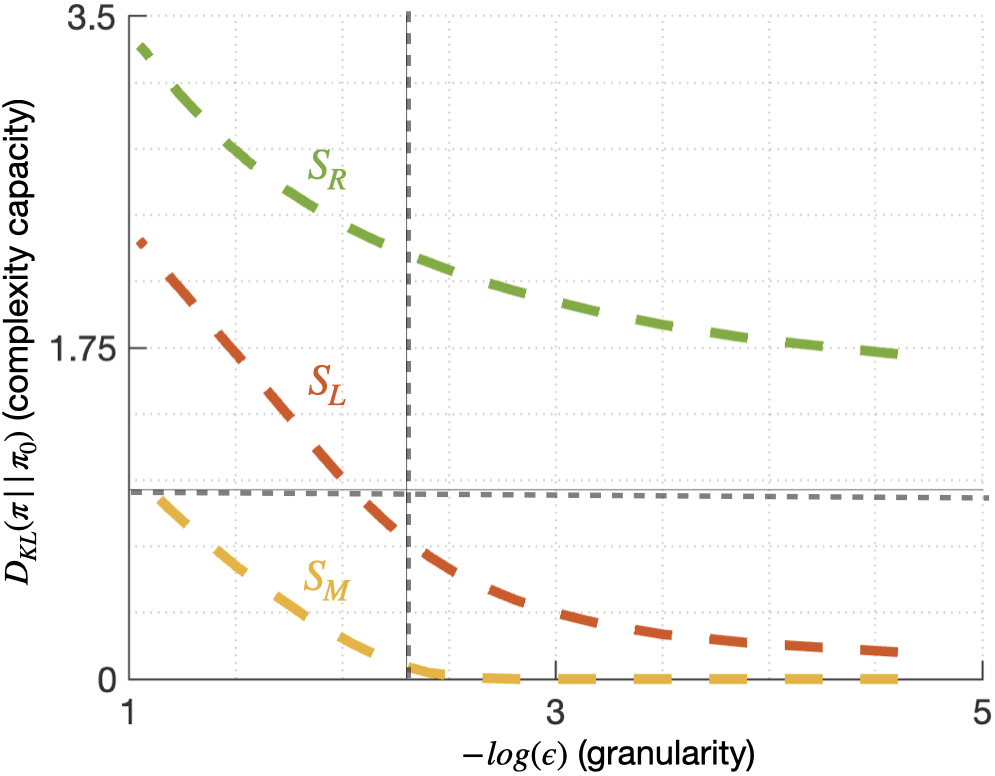} 
\caption
{\small \textbf{Complexity-granularity curves:} Each line shows the policy complexity capacity, relative to the default policy $\pi_0(0,1)$ (ordinate) required to reach the corresponding telic state at a given representational granularity level, quantified by the negative log of the sensitivity parameter $\epsilon$ (abscissa). Dashed gray lines show the values used in the dual-goal navigation example: $\delta=1$ (horizontal) and $\epsilon=0.1$ (vertical)} 
\label{fig:gc_curves}
\end{figure}

\begin{figure}[ht]
    \centering
    \includegraphics[width=0.45\textwidth]{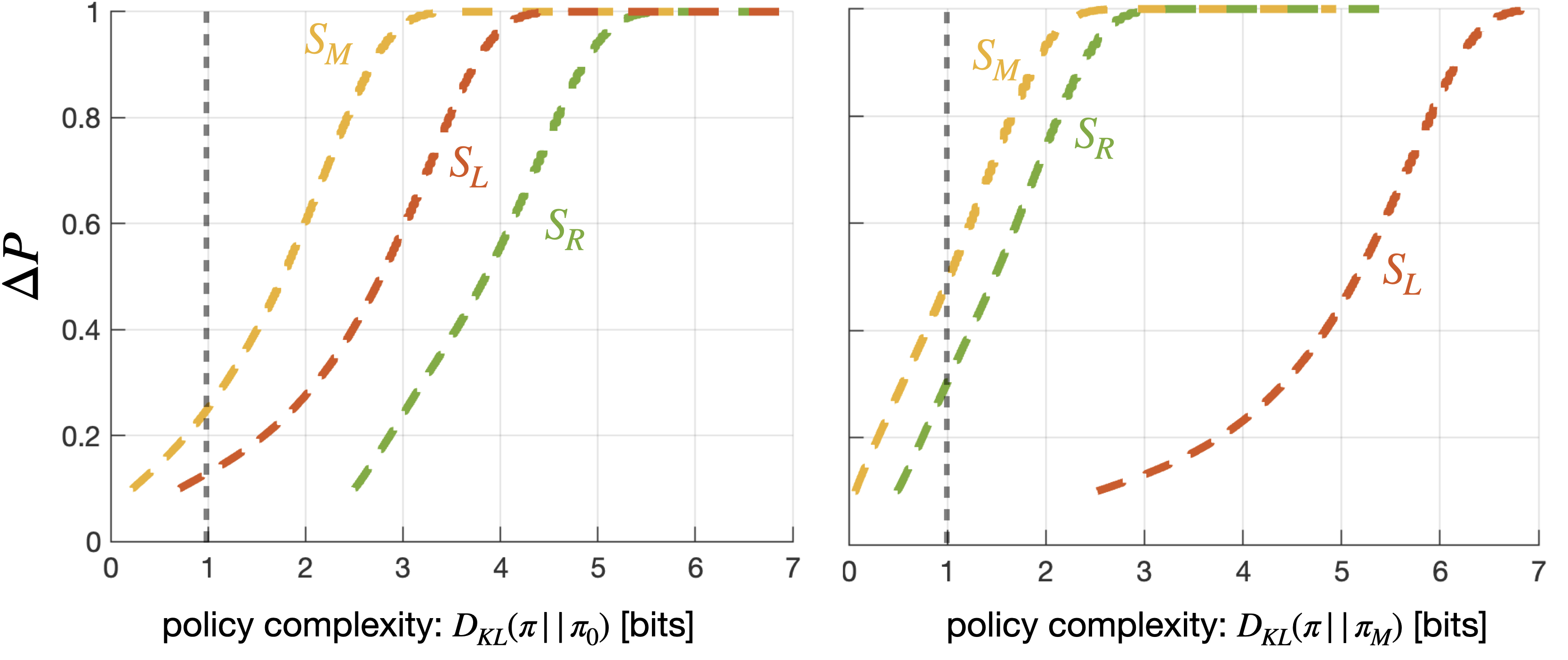}
    \caption{\small \textbf{Goal-complexity tradeoff curves:} the probability of reaching each telic state as a function of policy complexity. \textbf{Left:} an agent with a default policy $\pi_0:(\mu_0,\sigma_0)=(0,1)$ is unable to reach telic state $S_R$ with a complexity capacity limit of $\delta=1$ (gray vertical line). \textbf{Right:} with $\pi_1:(\mu_1,\sigma_1)=(1.09,1.24)$ as its default policy, the agent can reach both $S_M$ and $S_R$ with the same policy complexity capacity.}
    \label{fig:dp_complexity_curves}
\end{figure}

Using Eqs.~\ref{eq:random_walk_final_loc_dist} and~\ref{eq:telic_states_def} we can express each telic state in closed form, for example $S_R$ can be expressed, using the standard error function, $\operatorname {erf}{(x)}=2/\sqrt{\pi}\int^x_{0} e^{-t^2}dt$, as follows:
\begin{equation*}
\label{eq:telic_state_erf}
\begin{split}
&S_R=\{(\mu,\sigma)\bigg|\frac{1}{2}\left(\operatorname {erf} {\frac {R_{1}-T\mu }{{\sqrt {2T}}\sigma }}-\operatorname {erf} {\frac {R_{2}-T\mu }{{\sqrt {2T}}\sigma }}\right)-\\
&\frac{1}{2}\left(\operatorname {erf} {\frac {L_{1}-T\mu }{{\sqrt {2T}}\sigma }}-\operatorname {erf} {\frac {L_{2}-T\mu }{{\sqrt {2T}}\sigma }}\right)\geq \epsilon \}, 
\end{split}
\end{equation*}
with similar expressions for $S_L$ and $S_0$. 
To illustrate the notion of telic-controllability (Eq.~\ref{eq:telic_controllability}) using this representation, we define the complexity, $C(\pi)$, of a policy, $\pi(\mu,\sigma)$, with respect to the agent's default policy, $\pi_0(\mu_0,\sigma_0)$, as the KL divergence, per time step, between them: 
 \begin{equation*}
C(\pi)\equiv D_{KL}(\pi\|\pi_0).
 \label{eq:policy_complexity_dkl}
 \end{equation*}
The contour lines in the first three panels of Fig.~\ref{fig:phase_plots} (top \& bottom left) show isometric policy complexity levels for an agent with a complexity capacity of $\delta=1$ bit per time step, and a default policy $\pi_0(\mu_0=0,\sigma_0=1)$. Initially, both telic states, $S_R$, and $S_L$, lie within the range of the agent's policy complexity capacity (top left). The policies in $S_R$ and $S_L$ that are closest in the $KL$ sense to $\pi_0$ (green and red dots, respectively), both lie within a range of less than $\delta$ from $\pi_0$, i.e., the state representation is telic-controllable. When the center of $R$ shifts from $x_R=2$ to $x_R=2.5$ (top right), telic state $S_R$ is no longer within complexity range $\delta$ from $\pi_0$ and the state representation becomes non-controllable. To address this (bottom left), the state representation learning algorithm described in \ref{sec:learning_algorithm}, splits $S_R$ by adding an intermediate telic state $S_M$ (orange), centered around the policy closest to $S_R$ that is still within a KL-range of $\delta$ from $\pi_0$ (yellow dot). This changes the shape of $S_R$ and $S_L$ since now the probability of reaching each of the three telic states, $S_R,S_L$ and $S_M$, is defined in with respect to the two others, e.g., $S_M=\left\{(\mu,\sigma)|\Delta P_M(\mu,\sigma)\geq \epsilon\right\}$ where $\Delta P_M=p(x_T\in M\mid \mu,\sigma)-\max\{p(x_T\in L\mid \mu,\sigma),p(x_T\in R\mid \mu,\sigma)\}$, and similarly for $S_R$ and $S_L$. Since $\pi_M$ is, by construction, within a KL range of $\delta$ from $\pi_0$, the agent can reach $S_M$ by updating its default policy to $\pi_M$ (bottom right), bringing $S_R$ into reach again. Hence, the new state representation, consisting of $S_0,S_L,S_M$ and $S_R$, is tellic-controllable. Fig.~\ref{fig:dp_complexity_curves} illustrates the telic-complexity curves, showing the probability of reaching each telic state  achievable for a given complexity capacity level (x-axis). These curves quantify the maximal gain in the probability of reaching each telic state, $S_R,S_M$ or $S_L$, relative to the other two (ordinate), for a given policy complexity capacity level, with respect to a default policy of $\pi_0$ (left) or $\pi_M$ (right) (abscissa). 
Finally, Fig.~\ref{fig:gc_curves} illustrates the granularity-complexity tradeoff: the granularity of the state representation, quantified as $-\log(\epsilon)$ (abscissa), controls the complexity capacity required to reach each state (ordinate). Finer-grained representations are generally more controllable. For a granularity level of $\epsilon=0.1$ (gray vertical line), only $S_L$ and $S_M$ are reachable from $\pi_0(0,1)$ under a complexity capacity of $\delta=1$ (gray horizontal line).
\section{Discussion}
We illustrated a novel approach to modeling purposeful behavior in bounded agents, based on the hypothesis that goals, defined as preferences over experience distributions, play a fundamental role in shaping state representations. Coupling together descriptive and normative aspects of learning models, our framing posits a granularity-complexity tradeoff as a normative theoretical criterion guiding cognitive agents in determining which features of their environment to attend to and which to ignore in the context of a particular task \cite{niv2015reinforcement,langdon2019uncovering}. We accordingly hypothesize that goal selection can be usefully viewed as a processes of balancing two competing cognitive loads: representational granularity and policy complexity. The former limits the resolution of the goals (and the corresponding telic state representation) that the agent selects, while the latter controls the complexity of the policy generation, preventing behavior 
Clearly, our approach is highly simplified and entails theoretical assumptions which may do not hold in the general case. For example, computing the telic-distance, requires that the agent knows, or at least has a good model of the environment dynamics, which may not be the case for complex, real-world environments. While beyond the scope of this paper, these limitations could be potentially addressed using estimation and learning theoretic methods for bounding the telic-distance error under partially-observed approximations of the environment dynamics.   
While several methods for goal-directed state abstraction have been previously proposed \cite{li2006towards,shah2021value,kaelbling1993learning, zhang2020learning,steccanella2022state} our approach is different in suggesting that telic state representations are only defined with respect to a goal (rather than, for example, defining goals as a subset of preexisting states). Our approach is aligned with work using resource rational analysis to explain human learning and representation \cite{prystawski2022resource,lieder2020resource,ho2022people,correa2025exploring} but our emphasis here is on developing a principled theoretical account of how goals shape state representation learning in complexity constrained cognitive agents. Our quantification of policy complexity follows previous work applying information theoretic principles in reinforcement learning \cite{tishby2010information,rubin2012trading} and cognitive science \cite{amir2020value, lai2024human}. 
Notably, our complexity-granularity curves (Fig.~\ref{fig:gc_curves}) qualitatively resemble rate-distortion curves in information theory~\cite{cover1999elements}, suggesting a new interpretation of state representation learning via information theoretic lens~\cite{arumugam2021deciding,abel2019state}. Finally, the duality between goals and state representations characterizing our approach may help address the thorny problem of goal formation: where do goals come from in the first place? Specifically, goals may be selected based on the properties of the state representations they produce. Our framework thus suggests that bounded agents, who need to balance control over the environment with behavioral adaptability (cf.~\citeA{klyubin2005empowerment}), would do well to choose goals that produce telic-controllable state representations.

\section{Acknowledgments}
This work was supported by grant no. U01DA050647 from the National Institute on Drug Abuse. 

\bibliographystyle{apacite}

\setlength{\bibleftmargin}{.125in}
\setlength{\bibindent}{-\bibleftmargin}

\bibliography{CogSci_Template}


\end{document}